\documentclass{article}
\usepackage{amsfonts}
\usepackage{spconf,amsmath,graphicx,booktabs,hyperref}

\usepackage{array}

\usepackage{xcolor}

\newcolumntype{L}[1]{>{\raggedright\arraybackslash}p{#1}}

\title{EEG-Captioner: Grounding EEG-to-Language Generation in Fundamental Signal Properties}
\name{
Leyuan Jin$^{1}$, 
Jiahao Mei$^{1}$, 
Yixuan Yuan$^{2}$, 
Shimeng Li$^{1}$, 
Mengyue Wu$^{1}$, 
Yue Ding$^{2}$
\thanks{Corresponding authors: Mengyue Wu and Yue Ding.}
}

\address{
$^{1}$X-LANCE Lab, Shanghai Jiao Tong University, Shanghai, China\\
$^{2}$Shanghai Mental Health Center, Shanghai Jiao Tong University School of Medicine, Shanghai, China
}

\begin{document}
\ninept
\maketitle

\begin{abstract}
Electroencephalography (EEG)-to-language models offer a promising approach for translating 
complex neural signals into human-readable descriptions. However, generating plausible language does not necessarily demonstrate that a model has learned to interpret the underlying EEG signal. A direct way to assess such interpretation is to probe objectively measurable properties of the signal itself. We therefore construct an EEG-language dataset of 160,781 segments annotated with signal-level question answering (QA) pairs, where each question targets one of five property families and requires a quantitative comparison involving spectral, spatial, temporal, or lateral signal properties. Each segment is also paired with a feature-grounded EEG caption. We further propose a Dual-Domain EEG Encoder that jointly models temporal waveforms and spectral-spatial representations, and introduce \textbf{EEG-Captioner}, a unified EEG-to-language framework trained with both signal-level QA and EEG caption supervision. 
Experiments show that the proposed encoder achieves the highest overall macro-averaged balanced accuracy (Macro-BA) among the compared EEG encoders, while EEG-Captioner achieves a Macro-BA of 88.81\% on signal-level QA and improves caption factual consistency from 82.06\% to 83.46\% over caption-only language adaptation, demonstrating the complementary benefits of signal-level QA and caption supervision.

Demo page: \mbox{https://lyJinAoki.github.io/EEG-Captioner/}.
\end{abstract}


\begin{table*}[t]
\centering

\caption{Composition of the proposed EEG-language dataset.
Each EEG segment generates one caption and 25 candidate relational questions across
five task categories, with five questions per category.
The reported relational counts are the numbers retained after
numerical validity and signal-quality filtering.}
\label{tab:dataset_composition}

\small
\setlength{\tabcolsep}{4pt}
\renewcommand{\arraystretch}{1.1}

\begin{tabular}{llL{6.9cm}L{4.0cm}r}
\toprule

\textbf{Component} &
\textbf{Task} &
\textbf{Question} &
\textbf{Answer Space} &
\textbf{Valid Count} \\

\midrule

Signal-level QA
& Dominant Band
& Which frequency band has the highest power in a given time interval and spatial region?
& Delta, Theta, Alpha, Beta, Gamma
& 652,345 \\

& Strongest Region
& Which brain region has the highest power in a given frequency band and time interval?
& Frontal, Central, Parietal, Occipital, Temporal
& 6,879 \\

& Peak Period
& In which time interval---Early, Middle, or Last---does a given EEG feature reach its highest value?
& Early, Middle, Last
& 311,861 \\

& Temporal Pattern
& How does a given EEG feature change across the Early, Middle, and Last intervals?
& Increasing, Decreasing, Increase-then-Decrease, Decrease-then-Increase, Stable
& 491,189 \\

& Lateral Dominance
& Which hemisphere shows stronger activity in corresponding channels or brain regions?
& Left, Right, Approximately Symmetric
& 784,302 \\

\midrule

EEG Caption
& Free-form Caption
& Please describe the EEG signal in one coherent English caption.
& In the early interval, the EEG is dominated by ,,,
& 160,781 \\

\bottomrule
\end{tabular}

\end{table*}

\section{Introduction}
\label{sec}

Electroencephalography (EEG) provides a non-invasive and temporally precise measurement of neural activity. Interpreting EEG requires reasoning jointly about multiple signal characteristics, including temporal dynamics, spectral patterns, spatial distributions, and hemispheric asymmetries. Large language models (LLMs) offer a promising way to translate such complex signals into human-readable descriptions. However, generating plausible text does not necessarily imply that an EEG-to-language model has captured the underlying signal characteristics. Faithful EEG descriptions should therefore be grounded in measurable signal characteristics.

Recent EEG-to-language studies have made progress toward such grounding. E$^2$-LLM~\cite{maE^2LLMBridgingNeural2026} integrates pretrained EEG representations with an LLM for affective reasoning, but does not explicitly supervise the underlying signal properties. CerebraGloss~\cite{guCerebraGlossInstructionTuningLarge2025} instruction-tunes a vision-language model for fine-grained clinical EEG interpretation, focusing primarily on clinically relevant EEG characteristics. NeuroNarrator~\cite{wangNeuroNarratorGeneralistEEGtoText2026} incorporates spectro-spatial and temporal modeling and uses LLM-based scoring to assess signal properties in generated captions. Despite these advances, existing approaches still lack explicit and comprehensive supervision and direct evaluation of fundamental EEG properties, making it difficult to determine whether generated descriptions are systematically grounded in the underlying temporal, spectral, spatial, and lateral characteristics.
This need for explicit signal-level supervision also raises a question of whether the EEG representations used by language-based models preserve these signal properties. EEG representation learning has progressed from compact convolutional and Transformer architectures, such as EEGNet~\cite{lawhernEEGNetCompactConvolutional2018} and SleepTransformer~\cite{phanSleepTransformerAutomaticSleep2022}, to large-scale pretrained models including BIOT~\cite{yangBIOTBiosignalTransformer2023}, EEGPT~\cite{wangEEGPTPretrainedTransformer2024a}, LaBraM~\cite{jiangLargeBrainModel2024}, and CBraMod~\cite{wangCBraModCrissCrossBrain2025}. Although strong on downstream tasks, these models are not primarily designed to preserve the EEG properties needed for grounded language generation.

To address these limitations, we first construct an EEG-language dataset that provides explicit supervision and evaluation of fundamental EEG properties through signal-level question answering (QA) and feature-grounded captions. The QA annotations cover five property families across temporal, spectral, spatial, and lateral dimensions.
To better preserve the signal information required by these properties, we further propose a Dual-Domain EEG Encoder that jointly models raw temporal dynamics and spectral-spatial structure. Instead of serializing precomputed EEG features as text, the encoder maps raw EEG into compact learned tokens, enabling more efficient and scalable EEG-to-language generation. Building on the dataset and encoder, we introduce EEG-Captioner, a unified framework jointly trained with signal-level QA and caption supervision for grounded EEG caption generation.

Experiments show that the proposed Dual-Domain EEG Encoder achieves higher macro-averaged balanced accuracy (Macro-BA) than the compared encoders on signal-level QA, while joint QA-caption training further improves both QA performance and grounded caption generation over the single-supervision variants. The main contributions are:

\begin{itemize}
\item We construct an EEG-language dataset that pairs EEG segments with signal-level QA and feature-grounded captions, enabling explicit supervision and direct evaluation of fundamental EEG properties alongside caption generation.
\item We propose a Dual-Domain EEG Encoder that jointly models temporal waveforms and spectral-spatial representations, achieving stronger performance than existing EEG encoders on the proposed signal-level QA tasks.
\item We introduce EEG-Captioner, a unified EEG-to-language framework that jointly learns from signal-level QA and caption supervision, improving both fundamental EEG understanding and grounded caption generation.
\end{itemize}

\section{Method}
\label{sec}

\begin{figure}[!ht]
    \centering
    \includegraphics[width=0.80\columnwidth]{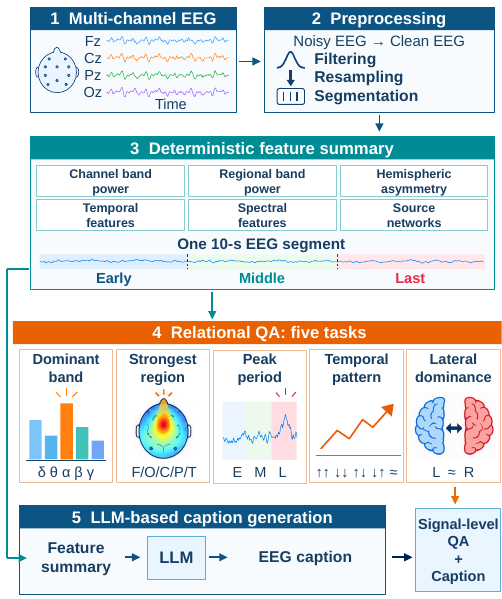}
    \caption{Illustration of the EEG-Language dataset construction pipeline.}
    \label{fig:pipeline}
\end{figure}

\begin{figure*}[t]
    \centering
    \includegraphics[width=\textwidth]{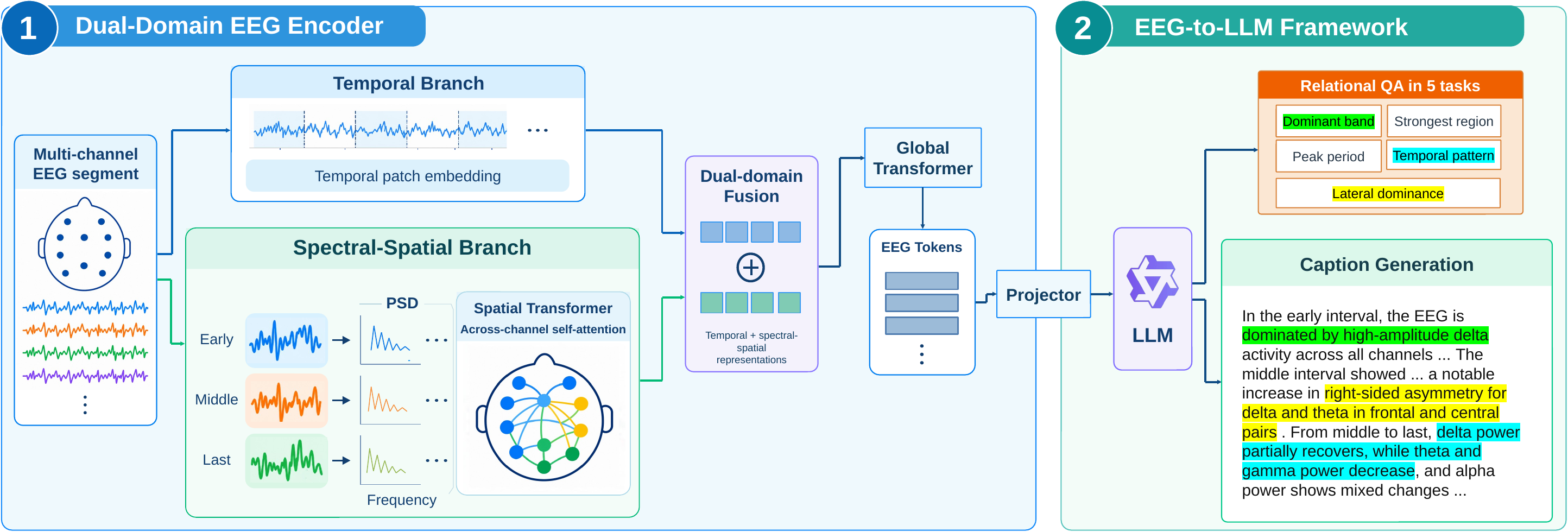}
    \caption{Overview of the proposed Dual-Domain EEG Encoder and EEG-to-LLM framework for relational QA and caption generation; representative outputs are shown on the right.}
    \label{fig:architecture}
\end{figure*}

The EEG-language dataset construction pipeline is illustrated in Fig.~\ref{fig:pipeline}, while the overall framework of EEG-Captioner is shown in Fig.~\ref{fig:architecture}. We first describe the construction of the EEG-language dataset, followed by the Dual-Domain EEG Encoder and the joint EEG-to-language training framework.

\subsection{EEG-Language Dataset Construction}
\label{sec:2.1}

We construct the EEG-language dataset from the Temple University EEG Corpus (TUEG) v2.0.2~\cite{obeidTempleUniversityHospital2016}. EEG recordings are processed using a standardized pipeline consisting of channel harmonization, filtering and resampling, fixed-length segmentation, and multi-level quality control.

For each valid EEG segment, we compute a deterministic numerical feature summary describing its temporal, spectral, spatial, and lateral characteristics. Specifically, each segment is divided into Early, Middle, and Last intervals, from which we extract channel-level band power, region-of-interest (ROI)-level band power, source-network power, inter-hemispheric asymmetry, time-domain features, and spectral-domain features.

Based on these features, we construct signal-level QA for five EEG property families: dominant band, strongest region, peak period, temporal pattern, and lateral dominance. Rather than regressing absolute feature values, which can vary substantially across recordings and subjects, we formulate the tasks as relational comparisons that capture interpretable structure across frequency bands, spatial regions, temporal intervals, and hemispheres. Each property family contains five predefined feature conditions, yielding 25 candidate QA pairs for each segment.
Ground-truth answers are computed directly from full-precision numerical features. We further apply task-specific validity and ambiguity criteria to remove unreliable annotations, including invalid feature values, insufficient separation between competing maxima, and inconsistent temporal patterns. The strongest region task retains fewer samples because regional comparisons require all constituent channels to satisfy the validity criteria.

Along with signal-level QA pairs, we generate a feature-grounded caption for each segment. The numerical feature summary is provided to DeepSeek-V4-Flash~\cite{deepseek-aiDeepSeekV4HighlyEfficient2026}, which is instructed to describe the observed temporal, spectral, spatial, and lateral characteristics while avoiding clinical or cognitive inference. Each sample therefore contains the raw EEG segment, the retained signal-level QA pairs, and an EEG caption, as summarized in Table~\ref{tab:dataset_composition}.

\subsection{Dual-Domain EEG Encoder}

We propose a \textbf{Dual-Domain EEG Encoder} to jointly represent information from the raw temporal waveform and its spectral-spatial structure. The encoder contains two parallel branches whose representations are subsequently aligned and fused.

In the temporal branch, each channel waveform is divided into short patches and projected into temporal tokens, allowing the encoder to model local waveform dynamics directly from the raw signal. In parallel, the spectral-spatial branch divides the EEG segment into Early, Middle, and Last intervals and computes channel-wise power spectral density (PSD) using Welch's method~\cite{welchUseFastFourier1967a}. Within each interval, channel self-attention~\cite{vaswaniAttentionAllYou2017} models dependencies across electrodes from the resulting spectral representations.

The outputs of the two branches are then aligned and fused to combine temporal dynamics with spectral-spatial context. Channel-wise attention pooling followed by a Global Transformer further aggregates information across channels and temporal positions, producing the final EEG tokens used by downstream language tasks.

\subsection{EEG-Captioner}

EEG-Captioner connects the Dual-Domain EEG Encoder to an LLM through a learnable projector. Given an EEG segment, the encoder produces a sequence of EEG tokens, which are mapped by the projector into the LLM embedding space and provided as multimodal context for language generation.

We jointly train the framework using the two annotation types provided by the EEG-language dataset. For signal-level QA, the model receives a property-specific question and generates the corresponding answer. For caption generation, the model generates a natural-language description of the signal characteristics. Both tasks share the same EEG encoder, projector, and LLM, while task-specific instructions specify the expected output format. The model is jointly optimized on both types of training examples.

\begin{table*}[t] 
\centering
\caption{
Comparison of EEG encoders on five relational tasks and EEG-input sanity checks.
In the Classifier stage, each EEG encoder is trained jointly with a relational classification head. In the Qwen stage, the trained encoder is frozen, and the projector and LoRA adapters are trained for relational instruction tuning. All values are balanced accuracy (\%), and Macro-BA is the unweighted mean
across the five task families.
}
\label{tab:table2}

\small
\setlength{\tabcolsep}{4.5pt}

\begin{tabular}{llcccccc}
\toprule
\textbf{EEG Encoder} &
\textbf{Stage} &
\textbf{Dominant} &
\textbf{Strongest} &
\textbf{Peak} &
\textbf{Temporal} &
\textbf{Lateral} &
\textbf{Macro-BA} \\
&
&
\textbf{Band(\%)} &
\textbf{Region(\%)} &
\textbf{Period(\%)} &
\textbf{Pattern(\%)} &
\textbf{Dominance(\%)} &
\textbf{(\%)} \\
\midrule

CBraMod & Classifier
& 71.68 & 72.59 & 34.13 & 33.60 & 40.27 & 50.45 \\

LaBraM & Classifier
& 90.56 & 70.60 & 89.87 & 71.68 & 57.60 & 76.06 \\

Dual-Domain & Classifier
& \textbf{93.76} & \textbf{91.29} & \textbf{93.87}
& \textbf{87.52} & \textbf{82.93} & \textbf{89.87} \\

\midrule

CBraMod & Qwen
& 61.76 & 69.53 & 36.27 & 36.00 & 36.27 & 47.96 \\

LaBraM & Qwen
& \textbf{89.76} & 72.23 & 84.00 & 68.80 & 52.53 & 73.47 \\

Dual-Domain & Qwen
& 88.64 & \textbf{87.01} & \textbf{94.67}
& \textbf{81.60} & \textbf{78.40} & \textbf{86.06} \\

\midrule
\multicolumn{8}{l}{
\textit{EEG input sanity check for Dual-Domain + Qwen}
} \\

Shuffled EEG & Qwen
& 20.00 & 20.37 & 35.73 & 18.72 & 28.53 & 24.67 \\

Zero EEG & Qwen
& 20.00 & 20.00 & 33.33 & 20.00 & 33.33 & 25.33 \\

\bottomrule
\end{tabular}
\end{table*}

\begin{table*}[t]

\centering
\caption{Comparison of EEG-Captioner training strategies.
Factual Consistency evaluates the accuracy of fundamental 
EEG signal descriptions in generated captions with respect 
to the corresponding numerical feature summary using GPT-5.6 Sol.}

\label{tab:table3}

\small

\setlength{\tabcolsep}{5pt}

\begin{tabular}{l|c|ccc}

\toprule

\textbf{Training} &
\multicolumn{1}{c|}{\textbf{Signal-level QA}} &
\multicolumn{3}{c}{\textbf{Caption Generation}} \\

\textbf{Supervision} &
\textbf{Macro-BA(\%) $\uparrow$} &
\textbf{Factual Consistency(\%) $\uparrow$} &
\textbf{ROUGE-L $\uparrow$} &
\textbf{BLEU-4 $\uparrow$} \\

\midrule

Caption-only adaptation

& 47.82

& 82.06

& 0.4722

& 0.3124 \\

Signal-level QA only

& 86.06

& 44.10

& 0.1470

& 0.0083 \\

Signal-level QA + Caption

& \textbf{88.81}

& \textbf{83.46}

& \textbf{0.4865}

& \textbf{0.3302} \\

\bottomrule

\end{tabular}

\end{table*}
\section{Experiments}
\label{sec:experiments}

\subsection{Experimental Setup}

\textbf{Dataset.} All experiments are conducted on the proposed EEG-language dataset.
The 160,781 EEG segments are split into 128,625 / 16,078 / 16,078
training, validation, and test samples in a filename-disjoint manner,
and the same partitions are used for all experiments.

\textbf{Evaluation metrics.}
For fundamental signal understanding, we report balanced 
accuracy (BA) for each relation family and Macro-BA, computed as their 
unweighted average. For caption generation, we report BLEU-4~\cite{papineniBleuMethodAutomatic2002} 
and ROUGE-L~\cite{linROUGEPackageAutomatic2004} to measure lexical overlap with reference captions. 
To assess signal grounding beyond lexical similarity, we further evaluate factual consistency using GPT-5.6 Sol through the 
ChatGPT web interface with High reasoning effort. The evaluator compares each generated caption with a feature summary derived from the corresponding EEG segment and assesses four signal-level aspects: dominant spectral pattern, temporal trend, spectral features, 
and spatial distribution, producing an overall factual consistency score from 0\% to 100\%.

\textbf{Implementation details.}
For the classifier stage, all encoders are trained with a single-layer classifier for 30,000 steps using AdamW, cosine decay, 1,000 warmup steps, and weight decay 0.01. CBraMod and LaBraM are initialized from pretrained weights and optimized with an encoder learning rate of $3\times10^{-5}$, whereas the Dual-Domain encoder is trained from scratch with a learning rate of $3\times10^{-4}$; all classifier heads use a learning rate of $3\times10^{-4}$. Effective batch sizes are 16 for CBraMod/LaBraM and 48 for Dual-Domain.
For the encoder-to-Qwen stage, the encoders from the classifier stage are frozen and connected to frozen Qwen2.5-Omni-7B~\cite{xuQwen25OmniTechnicalReport2025} through a newly initialized projector. Low-Rank Adaptation (LoRA)~\cite{huLoRALowRankAdaptation2021} is applied to the Q/K/V/O projections of the language model ($r=8$, $\alpha=32$, dropout=0.05). The projector and LoRA parameters are trained for 30,000 steps using AdamW with a learning rate of $1\times10^{-4}$, weight decay 0.01, and an effective batch size of 4 on two GPUs.
For EEG-Captioner, all variants use the same frozen Dual-Domain EEG encoder obtained from the classifier stage, thereby isolating differences in the language-adaptation strategy from differences in EEG-encoder pretraining. For the caption-only adaptation, the projector and LoRA parameters are re-initialized and trained for 15,000 steps using only caption supervision at a learning rate of $5\times10^{-5}$. Validation performance had converged by the end of the 15,000-step training. EEG-Captioner instead initializes the projector and LoRA parameters from the signal-level QA checkpoint and continues training for three consecutive 15,000-step stages with caption fractions of 0.25, 0.75, and 0.90, respectively. All captioning stages use AdamW with weight decay 0.01, an effective batch size of 4, and seed 42.

\subsection{Evaluation Protocols}
\textbf{Encoder comparison.}
We compare the proposed Dual-Domain EEG Encoder with LaBraM and CBraMod, two representative state-of-the-art EEG encoders, at both representation and language levels. At the representation level, performance is evaluated over the five signal-level QA property families using balanced accuracy. At the language level, all encoders are evaluated with the same Qwen-based setup on signal-level QA. We additionally assess the dependence of the Dual-Domain + Qwen model on EEG input through perturbation experiments, replacing the original EEG with cross-sample shuffled EEG or zero signals while keeping the questions unchanged.

\textbf{EEG-Captioner evaluation.}
We compare EEG-Captioner with two variants under the same model configuration. The caption-only adaptation variant trains only the language adaptation modules with caption supervision, while retaining the same frozen QA-pretrained Dual-Domain encoder used in the other variants for a controlled comparison. The signal-level-QA-only variant corresponds to the model after signal-level QA training, without subsequent caption training. Captioning performance is evaluated on the same held-out test set using identical metrics across all variants.
\section{Results}
\label{sec:results}

\noindent\textbf{Comparison of EEG Encoders.}
Table~\ref{tab:table2} compares different EEG encoders under signal-level QA evaluation. At the classifier stage, the proposed Dual-Domain EEG Encoder achieves 89.87\% Macro-BA, outperforming LaBraM (76.06\%) and CBraMod (50.45\%). The gains are particularly pronounced for strongest-region, temporal-pattern, and lateral-dominance prediction, suggesting that the proposed representation better preserves information relevant to these signal-level EEG properties.

After transferring the frozen encoders to Qwen, the Dual-Domain encoder retains its advantage, achieving 86.06\% Macro-BA compared with 73.47\% for LaBraM and 47.96\% for CBraMod, suggesting that the signal-level information captured by the proposed encoder can be effectively transferred to language-based QA.

EEG input perturbation further leads to substantial performance degradation, suggesting that the model’s predictions depend on the EEG input rather than solely on language-side shortcuts.

\noindent\textbf{Effectiveness of EEG-Captioner.}
Table~\ref{tab:table3} compares the three variants using the same frozen QA-pretrained EEG encoder. On signal-level QA, caption-only adaptation achieves 47.82\% Macro-BA, while signal-level-QA-only training reaches 86.06\%. EEG-Captioner further improves Macro-BA to 88.81\%, suggesting that caption supervision can complement signal-level QA training.
For caption generation, EEG-Captioner improves factual consistency from 82.06\% to 83.46\% compared with caption-only adaptation, while increasing ROUGE-L from 0.4722 to 0.4865 and BLEU-4 from 0.3124 to 0.3302. Overall, the joint training strategy maintains strong signal-level QA performance while providing modest improvements in caption quality.
\section{Conclusion}
\label{sec:conclusion}

In this paper, we present EEG-Captioner, a unified EEG-to-language framework for grounded EEG understanding. We construct an EEG-language dataset that pairs EEG captions with signal-level QA pairs covering fundamental temporal, spectral, spatial, and lateral signal properties. We further introduce a Dual-Domain EEG Encoder to model temporal and spectral-spatial information. Experiments show that the proposed encoder achieves improved performance on signal-level QA tasks, while joint QA-caption training improves both signal-level QA performance and caption factual consistency. These results suggest that incorporating explicit signal-level supervision can help improve the grounding of EEG-to-language generation in measurable EEG properties.

\section{Compliance with Ethical Standards}

This study used the de-identified Temple University Hospital EEG Corpus (TUEG), which was collected with approval from the Temple University Institutional Review Board and in accordance with the Declaration of Helsinki~\cite{obeidTempleUniversityHospital2016}. No new participants were recruited or data collected in this study.

\bibliographystyle{IEEEbib}
\bibliography{references}

\end{document}